# Deep Learning-based Quality Assessment of Clinical Protocol Adherence in Fetal Ultrasound Dating Scans

Sevim Cengiz*
Sevim.cengiz@mbzuaia.ac.ae
Mohammad Yaqub*
Mohammad.yaqub@mbzuai.ac.ae

*Department of Computer Vision, Mohamed bin Zayed University of Artificial Intelligence, Abu Dhabi, United Arab Emirates

## Abstract

To assess fetal health during pregnancy, doctors use the gestational age (GA) calculation based on the Crown Rump Length (CRL) measurement in order to check for fetal size and growth trajectory. However, GA estimation based on CRL, requires proper positioning of calipers on the fetal crown and rump view, which is not always an easy plane to find, especially for an inexperienced sonographer. Finding a slightly oblique view from the true CRL view could lead to a different CRL value and therefore incorrect estimation of GA. This study presents an AI-based method for a quality assessment of the CRL view by verifying 7 clinical scoring criteria that are used to verify the correctness of the acquired plane. We show how our proposed solution achieves high accuracy on the majority of the scoring criteria when compared to an expert. We also show that if such scoring system is used, it helps identify poorly acquired images accurately and hence may help sonographers acquire better images which could potentially lead to a better assessment of conditions such as Intrauterine Growth Restriction (IUGR).

## 1. Introduction

Fetal size and growth trajectories are the key indicators of fetal health and the detection of possible fetal abnormalities (Mayer and Joseph,2013). Abnormal growth is commonly described as small for gestational age (SGA) and large for gestational age (LGA) in the medical literature (Damhuis et al.,2021). A fetus that has SGA has a higher risk of peri- natal morbidity and death (Trivedi and Puri,2015). In low- and middle-income countries (LMICs), 32.4 million infants were born with SGA in 2010, accounting for 27% of all live births (Black,2015). Additionally, perinatal mortality has been reported at a higher rate in LMICs throughout the world (Zupan,2005). Early detection of abnormalities with ges- tational development might be used to predict perinatal death and morbidity (Alexander et al.,1995) (Callaghan and Dietz,2010). The most reliable way to accurately estimate gestational age (GA) is through an early ultrasound exam (commonly known as the Dating scan) that measures the fetus between 8 and 13 weeks of pregnancy rather than a later ultrasound examination (Trivedi and Puri,2015) (Peleg et al.,1998). Correct measurement of Crown-rump length (CRL) is an extremely important biometric measurement starting from the top of the fetus's head to the bottom of the rump, in order to achieve an accurate estimation of GA. As a result, an accurate CRL measurement is extremely important to detect possible fetal abnormalities such as SGA, and LGA.

To ensure that the CRL measurement is clinically representative, the plane where this measurement is performed has to meet a few clinical guidelines. Acquiring an image of the fetus that meets these guidelines is not an easy task even for senior sonographers in some cases. This is due to different reasons including the complexity of the clinical protocol, fetal movement, variable maternal characteristics, sonographer's experience and the ultrasound machine at hand. Therefore, we believe that quality assessment of CRL measurement plays a valuable role in the reliability of clinical results.

The National Health Services Fetal Screening Programme (NHS-FASP) in the United Kingdom proposed a guideline for CRL and NT measurement to improve fetal images' consistency and reproducibility as a standard method (NHS Fetal Screening Programme, 2012a). The guidelines check for image magnification, fetal position and fetal attitude, linear CRL measurement, and caliper placement of the CRL measurement (NHS Fetal Screening Programme,2012b). A study by Wanyonyi et. al used saved CRL images from each participant with a first trimester ultrasound, and the quality of the images are judged by the radiologist using a criteria set based on a image-scoring system (Wanyonyi et al., 2014). The study was conducted to compare an objective evaluation of two independent sonographers comprised of criterion-based CRL measurement. If the requirement was not fulfilled, a score of zero was assigned for each component on the image score. The horizontal orientation (97.6%) and good magnification (95.9%) had the highest inter-reviewer agree- ment, whereas the neutral position of the picture and mid-sagittal section had the lowest (78.2% and 81.5%, respectively). This study showed how the results were dependent on the scoring of the sonographers. Therefore, there is a need for an objective image-scoring system to decrease the effects of the different image scoring approaches based on the sonographers. AI-driven algorithms have the potential to solve such problems by creating a reproducible and accurate solutions.

Automated computer-based approaches are becoming more popular among researchers as deep learning technologies for medical image assessment progress. Using image charac- teristics and a regression neural network, an AI-based estimation of fetal GA was recently done (Bradburn et al.,2020a). Another research looked at how to identify fetal imaging planes on prenatal ultrasonography throughout different GAs (Bradburn et al.,2020b). The authors of (W-lodarczyk et al.,2019) claimed that the machine learning approach of their study may predict premature birth. The approach they introduced is based on the segments of cervical length (CL) and anterior cervical angle (ACA) and utilizes these esti- mates to classify premature births. A follow-up study used a multi-tasking U-Net network to segment the cervix, improving the prediction of preterm birth (W-lodarczyk et al.,2020). Another AI-based study used convolutional regression networks to do an automatic brain maturation estimation from 3D ultrasound images (Namburete et al.,2017).

The main goal of this research study was to develop an automatic AI-based fetal image- scoring system to evaluate the adherence of acquired CRL image in Dating scans to the clinical guidelines which should help improve the accuracy of CRL measurement and con- sequently fetal GA estimation. Our proposed method depends on precisely segmenting the fetal head and body in order to reliably verify the clinical criteria in the CRL image, which is subsequently utilized to determine fetal GA. We present a rigorous examination of the devised approach and compare the two different manual (human-driven) and automatic (computerized) image-scoring systems. To out knowledge, this is the first attempt to tackle the problem of automatic quality assessment in Dating scans.

## 2. Materials and Methods

### 2.1. Dataset

695 fetal ultrasound images from the first trimester scan were extracted from a hospital archive. All ultrasound images were from normal pregnancies excluding twin pregnancies. All images were anonymized before any use.

All fetal ultrasound images were manually segmented into 3 classes (head, body, and fetal palate) by an expert using ITK-Snap drawing annotation tool. If the fetal palate was not visible, the images were segmented into 2 classes (head and body). The manual segmen- tation of images was reviewed by another expert for additional correction if needed. Four hundred ninety-eight images were randomly selected and used to train the segmentation model and 197 images were used to test and evaluate the segmentation accuracy.

### 2.2. Fetal Head, Body, Palate Segmentation

In a variety of applications, U-Net (Ronneberger et al.,2015) has shown to be effective for segmenting objects in images. The basic U-Net network comprises a four-block fully convolutional encoder-decoder architecture, each of which is made up of several convolu- tional and ReLU layers, followed by a max-pooling layer. For multi-class segmentation, we investigated multiple hyper-parameters such as image size, network depth, the number of epochs and learning rate. The number of epochs was 100 and the learning rate was set to 1e-5. The Tensorflow library was used to implement and evaluate the proposed method.

### 2.3. Data Augmentation

Several data augmentation methods were performed to establish more accurate and robust segmentation because of the limited dataset. Horizontal and vertical flip, brightness, con- trast, and rotation (degrees [±10]) were applied. We only performed data augmentation during the training stage. There was no data augmentation done during the validation stage.

### 2.4. Assessment of fetal CRL image scoring criteria

The image scoring criteria based on the FASP ultrasound guidelines (NHS Fetal Screen- ing Programme,2012b) and the paper (Wanyonyi et al.,2014) were defined to create a set of criteria for acquiring the optimal CRL plane and measuring the CRL. The guidelines suggest that the head of the fetus should be in a neutral position, with no hyperflexion or hyperextension. The midline-sagittal section and the fetal palate should be visible. The crown and rump should be clearly defined for the best caliper placement and the longest measurement of the fetus. CRL axis should be parallel to the horizontal line or the angle of the CRL axis and horizontal line should be between ±15 degrees. The entire CRL section of the fetus should fill over %60 of the fetal ultrasound image. A spreadsheet included each criterion to evaluate the adherence of acquired CRL image in Dating scans to the clinical guidelines and this table was filled according to the manual (human-driven) and automatic (computerized) image-scoring systems. Later, this spreadsheet was used for comparison between two different systems, manual (human-driven) and automatic (com- puterized) image-scoring systems. We will now discuss the method we developed to handle each one of these 7 criteria.

**Fetal structure segmentation.** U-Net was trained with 498 images. One hundred ninety seven fetal ultrasound images were segmented using the model weights and segmented images were saved for further evaluation.

**Mid-sagittal section**. Predicted segmentation images were scanned to find the 1,2, and 3 label numbers that represent the head, body and, fetal palate, respectively. In this step, the existence of label 3 in the segmented images defined whether there was a fetal palate in the image or not. The existence of the fetal palate was used to understand that the fetus was screened at the mid-sagittal section.

**CRL measurement**. CRL measurement was calculated based on the method described in our previous study (Cengiz and Yaqub,2021). Briefly, we calculated the CRL measure- ment as the longest distance between contour points using the segmentation mask. The difference between the mask image and a dilated mask with a kernel size of 3x3 was used to generate the segmentation contour. The head and body segmentation contours were merged as a one-class. Then, the longest distance between two points was computed as the Euclidean distance. The maximum Euclidean distance was used to measure the CRL in the pixel space (Figure 2- Blue line between A and B point).

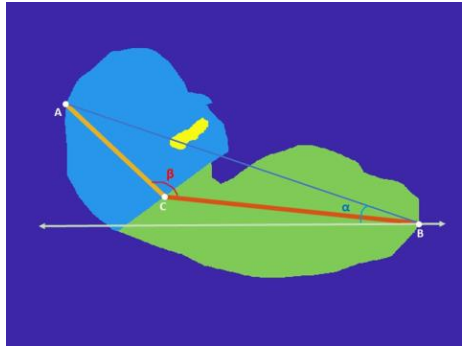

**Figure 1:** An example of a segmented image showing CRL measurement between A and B points (blue line), middle point (C) between the head and the body border, alpha representing the angle between the CRL line and the horizontal line, and beta representing the angle between the head and the body.

**Horizontal orientation**. The top of the head (Figure 1, point A) and the bottom of the body (Figure 1, point B) were used to calculate the equation of the CRL line. Then, the slope was calculated between the CRL line passing through the given points and the horizontal axis line. This angle was used to verify the criteria that is CRL axis should be parallel to the horizontal line or the angle of the CRL axis and horizontal line should be between ± 15 degrees. If the angle is between ± 15 degrees, the image criteria score for horizontal orientation was tagged as 1.

**Magnification**. The magnification criterion of the fetus was determined with whether the horizontal projection length of the CRL line, which was parallel to the x-axis, was higher from the %60 of the image.

**Neutral position**. The borderline between the head and body was established after finding the neighbors points between the contour points of the segmented head and body image. The middle point of the borderline was used to calculate the angle between head and body. The law of the Cosines was used the find the angle between the line passing through between the point A (Figure 1) and the middle point (Figure 1, point C) and the line passing through between the point B (Figure 1) and the middle point (Figure 1, point C). This angle was used to classify fetal image position as a normal, or non-normal position such as hyperflexion, or hyperextension. First, the segmentation performance, middle point between head and body, crown and rump coordinates, and neutral position of all images were examined and then the images which had poor segmentation, wrong crown, rump and middle points determination, and non-normal position were excluded. The rest of the images were used to calculate for the mean and standard deviation (SD) of the angles between head and body (153.81±9.17). Based on the mean and SD of the normal angle, we determined the range which represents normal (neutral) position of the fetus as between
144.64 and 162.88 and the images between the range were tagged as 1 and the rest were tagged as 0.

**Fetal face up/down**. Whether the face of the fetus looking up or down in the womb was determined the point (Figure 2, red circle) place in the quadrant coordinate by lining a perpendicular line from the middle point between head and body to the CRL line. If the point places at the +/- Y axis of the quadrant coordinate, it was tagged as the face of the fetus looking up or down, respectively.

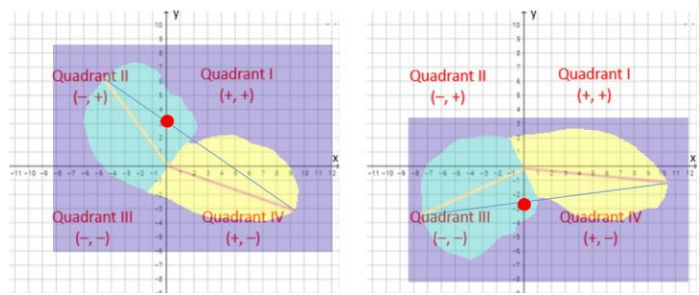

**Figure 2:** Examples of segmented images, which face was up (left) and down (down) at the quadrant coordinate.

**Visibility for correct left and right caliper placements**. The regions of the left and right side of the fetus were analyzed by creating a 10x5 size square from the top of the head and bottom of the body points to understand that these points were clearly visible. For this purpose, we evaluated how many numbers of pixels have the 0 pixel intensity, which represents the black area. If the number was over the half of pixels in the square, we tagged the point as clearly visible and it was a good point for caliper placement.

## 2.5. Statistical analysis

The images were evaluated to create two different image scoring tables including 7 items listed above by an AI-based approach and an expert-based. Images were tagged as accept- able if the score number out of 7 criteria was more than and equal to 4. An expert-based table was accepted as a ground truth (GT) and the AI-based table was used as an observed cases. Then, confusion matrix was created between the GT and observed cases to calcu- late the accuracy, precision and recall scores. An inter-rater reliability analysis using the Cohen's weighted kappa statistic was performed to determine the consistency among two image-scoring approaches. The acceptance criteria for CRL measurement, which included seven components, were evaluated to a reliability analysis by carrying out Cronbach's alpha. IBM SPSS Statistics was used for statistical analysis.

## 3. Results

Accuracy, precision and recall scores are listed in Table 1. Horizontal orientation, magnifi- cation, and fetal face position have the highest accuracy scores. Additionally, magnification, horizontal orientation and fetal face position have the highest precision scores compared to the precision scores of the neutral position, and fetal palate (mid-sagittal section). On the AI-based image scoring approach, 148 out of 197 images are found acceptable. On the sub- jective image scoring approach, 170 out of 197 are found acceptable. The overall accuracy of the proposed method for the adherence and acceptability of the images is 74.6%.

Table 2 shows the p-value and Cohen's weighted kappa with 95 % confidence intervals for inter-rater agreement for categorical scales. Firstly, the inter-rater agreement of acceptance of the CRL measurement between AI-based and expert-based image scoring approaches is kappa = 0.201 (p = 0.003), 95% CI (0.050, 0.352). Secondly, each criterion in the both scoring systems is evaluated individually for the agreement. There is substantial agreement of the magnification between the two image scoring approaches (k = 0.721, p = 0.000). However, we show a fair agreement between the two reviewers considering fetal palate, left and right definition for caliper placement (k = 0.313, 0.352, 0.315, respectively, and p
<0.001 for all).

Table 3 demonstrates the reliability analysis, which is carried out on the acceptance criteria of CRL measurement of expert- and AI-based image scoring comprising 7 items. Cronbach's alpha is $\alpha = 0.340$ and $\alpha = 0.464$, which indicates an unacceptable level of internal consistency items for expert- and AI-based image assessment with this specific sample. The two exceptions for the CRL measurement acceptance are criteria neutral position and right definition for caliper, which would increase the Cronbach's alpha to $\alpha = 0.351$ and 0.382 for the expert- image scoring approach for acceptance of CRL measurement. The exception of criterion horizontal orientation would increase to Cronbach's alpha to $\alpha = 0.501$.

**Table 1:** Accuracy, precision and recall scores of criteria for CRL measurement

| Criteria | Accuracy (%) | Precision (%) | Recall (%) |
|---|---|---|---|
| 1-Neutral position | 53.8 | 53.0 | 100 |
| 2-Horizontal orientation | 90.9 | 90.9 | 100 |
| 3-Fetal Palate (Mid-sagittal section) | 83.2 | 57.9 | 30.6 |
| 4-Magnification | 90.4 | 98.0 | 90 |
| 5-Left definition for caliper | 76.1 | 76.1 | 100 |
| 6-Right definition for caliper | 72.1 | 72.1 | 100 |
| 7-Fetal face up/down | 90.9 | 90.9 | 100 |
| 8-Acceptance of CRL measurement | 74.6 | 90.5 | 78.8 |

**Table 2:** For each individual criterion of image rating for CRL measurement, adjusted kappa and percentage of agreement were calculated. *LB = lower bound, *UP = upper bound.

| Criteria | p | Cohen's weighted kappa (95% CI [LB*,UP*]) |
|---|---|---|
| 1-Neutral position | 0.727 | -0.024 [-0.162, 0.113] |
| 2-Horizontal orientation | 0.767 | -0.019 [-0.140, 0.102] |
| 3-Fetal Palate (Mid-sagittal section) | <0.001 | 0.313 [0.139, 0.487] |
| 4-Magnification | 0.000 | 0.721 [0.605, 0.837] |
| 5-Left definition for caliper | <0.001 | 0.352 [0.244, 0.460] |
| 6-Right definition for caliper | <0.001 | 0.315 [0.204, 0.427] |
| 7-Fetal face up/down | 0.003 | 0.209 [-0.008, 0.426] |
| 8-Acceptance of the CRL measurement | 0.003 | 0.201 [0.050, 0.352] |

**Table 3:** Inter-item reliability test for the CRL measurement acceptance of expert-based and AI-based image scoring.

| Criteria | $\alpha$ for excluding each individually | |
|---|---|---|
| | **Expert-based** | **AI-based** |
| 1-Neutral position | 0.351 | 0.414 |
| 2-Horizontal orientation | 0.296 | 0.501 |
| 3-Fetal Palate (Mid-sagittal section) | 0.244 | 0.430 |
| 4-Magnification | 0.282 | 0.339 |
| 5-Left definition for caliper | 0.261 | 0.462 |
| 6-Right definition for caliper | 0.382 | 0.437 |
| 7-Fetal face up/down | 0.316 | 0.378 |
| Alpha coefficient for all seven items | 0.340 | 0.464 |

## 4. Discussion and Conclusion

We developed a deep learning approach which assesses the quality adherence of fetal CRL images in Dating scans to clinical guidelines. This is important to ensure that the per- formed CRL measurement on that view is clinically correct. In this paper, segmented fetal structures were used to guide the assessment of the criteria of the CRL view. The magnification, horizontal orientation, and face position were detected well. However, the neutral position which was evaluated considering the angle between head and body was a challeng- ing part. Also determining the mid-sagittal section considering the existence of the fetal palate segmentation was also a challenging part of the study.

There was a good accuracy rate for the horizontal orientation, magnification, fetal face position. Even if the fetal palate accuracy was high, the precision and recall scores were low due to the limited fetal ultrasound images that had fetal palate. As a result, we had a higher number of true negatives and a limited number of false positives and false negatives, which lead to a higher accuracy rate but low precision and recall. The neutral position had the lowest accuracy and precision scores. We acknowledged the fact that the angle range of the neutral position we set in this study was arbitrary, and based on the min and max angle scores of the normal position of the fetal ultrasound images. There was a substantial level of inter-rater agreement for magnification between the expert-based and AI-based approaches (k = 0.721). However, there was no agreement between the neutral position and horizontal orientation according to Cohen's kappa. Due to the limited dataset that we had caused a low level of agreement between two image scoring approaches. The angle range between the head and body for the neutral position determination might be evaluated with an additional dataset, which will help to improve the results.

The limited dataset size is one of the study's drawbacks and the imbalanced dataset caused positives to be minor in the false negatives and there were a lot of positives that were in the true positives. This might indicate that the results we provide are not guaranteed if the approach is used to different datasets. In the future, more evaluation will be required. Furthermore, because all of the pictures we utilize come from the same ultrasound ma- chine (GE Voluson E8), estimating GA from images taken on different ultrasound machines requires further investigation.

Accurate CRL measurement plays an important in detecting fetal abnormalities such as LGA and SGA. The nuchal-translucency scan (NT) that is used to detect chromosomal abnormalities is done in the first trimester which focuses on a similar screening protocol as the one used for CRL. A study suggested that the measurement of crown-rump length should be taken into consideration when calculating the likelihood ratio for a particular NT in the first trimester screening because nuchal-translucency thickness increases with fetal crown-rump length during 10–14 weeks of pregnancy(Snijders et al.,1998). We believe that an AI-based automatic image-scoring approach for CRL measurement will also help improve the accuracy of the NT measurement considering the appropriate CRL position.